\documentclass{article}
\usepackage[final]{neurips_2022}
\usepackage{graphicx}
\usepackage[utf8]{inputenc} 
\usepackage[T1]{fontenc}    
\usepackage{hyperref}       
\usepackage{url}            
\usepackage{booktabs}       
\usepackage{amsfonts}       
\usepackage{nicefrac}       
\usepackage{microtype}      
\usepackage{xcolor}         
\usepackage{natbib}
\title{On Extending Semantic Abstraction for Efficient Search of Hidden Objects}

\author{%
  Nikhilesh Belulkar \\
  \texttt{nb2953@columbia.edu } \\
  Department of Computer Science \\
  Columbia University \\
  New York, NY 10025 \\
  \And
  Tasha Pais \\
\texttt{tdp2129@columbia.edu} \\
  Department of Computer Science \\
  Columbia University \\
  New York, NY 10025 \\
}
\begin{document}

\maketitle

\begin{abstract}
\textit{Semantic Abstraction}'s key observation is that 2D VLMs' relevancy activations roughly correspond to their confidence of whether and where an object is in the scene. Thus, relevancy maps are treated as "abstract object" representations. We use this framework for learning 3D localization and completion for the exclusive domain of hidden objects, defined as objects that cannot be directly identified by a VLM because they are at least partially occluded. This process of localizing hidden objects is a form of unstructured search that can be performed more efficiently using historical data of where an object is frequently placed. Our model can accurately identify the complete 3D location of a hidden object on the first try significantly faster than a naive random search. These extensions to semantic abstraction hope to provide household robots with the skills necessary to save time and effort when looking for lost objects.

\textbf{Additional Resources:} 
\href{https://github.com/tashapais/semantic-abstraction}{Code}

\end{abstract}
\section{Problem Definition}
In the world of personal assistant robots, retrieval of specific objects given a time constraint is an important task. For instance, a user could command a robot to "find my keys" while getting ready to leave the house. Since the robot has sufficiently explored the environment from past actions and has a quantitative understanding of where an individual's unique preference for an object's placement is, the robot's memory is much better suited to quickly search and retrieve an object while a user continues a different task. This can be a form of human-robot collaboration when the task is clear and often repeated. 

A point cloud is the most complete representation of an object in 3D space. When aggregating views of a partially hidden object from multiple camera angles, it can be difficult to localize the object. Given that an object may be hidden in one view and visible in another, we concatenate an object's point cloud from multiple views.
\section{Method}
Our method uses \href{https://doi.org/10.48550/arxiv.2207.11514}{semantic abstraction} as the primary system over which we develop our algorithm to optimize search.

The key steps of our procedure are summarized below:
\begin{enumerate}
\item Collect multiple egocentric views of the environment through interaction, ex: opening cabinets and drawers. 
\item From these views, query semantic abstraction with objects and their labels to generate point clouds for individual objects, ex: "apple in fridge", "tomato in fridge", "fork in drawer".
\item Aggregate the location data from several views and use the Expectation-Maximization algorithm to generate a Gaussian Mixture Model (GMM) about the location of a particular object. 
\item Once this GMM has been developed, use it to optimize search times when queried for a specific object.
\end{enumerate}

\subsection{Step One}
The robot interacts with its environment collecting several different views. Since this is an embodied AI task, we chose RoboTHOR as the simulation environment. This random interaction is 
currently an explicitly predefined procedure in our code. In the future, we intend to use an LLM like ChatGPT to translate natural language queries into actions and provide an understanding of where certain objects are typically located to begin search. 

\begin{figure}[htp]
    \centering
    \includegraphics[width=4cm]{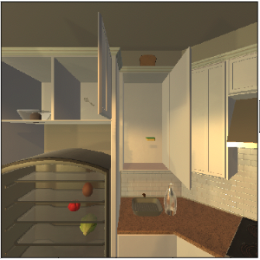}
    \caption{The camera view most often used to query Semantic Abstraction}
    \label{fig:galaxy}
\end{figure}

\subsection{Step Two}
Given an image, we utilized much of the infrastructure provided by \href{https://doi.org/10.48550/arxiv.2207.11514}{semantic abstraction} to generate a point cloud representing the relevancy of different parts of the image to the query provided. 

A query is made to CLIP. CLIP is a database that contains images and captions. Through contrastive learning CLIP localizes the parts of an image relevant to a caption. It utilizes this understanding to score how relevant each pixel of an image is to a particular query. For instance in the image below a relevancy score is provided to each pixel in the image when queried with "fork in cabinet".

Further, we also used the intrinsic and extrinsic camera parameters of each view to map where each pixel is in three-dimensional space. This is best visualized through the point cloud produced below.

\begin{figure}[htp]
    \centering
    \includegraphics[width=4cm]{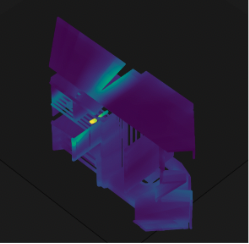}
    \caption{Query to localize the location of a fork inside a cabinet}
    \label{fig:galaxy}
\end{figure}

Once each pixel's 3D location is mapped to its corresponding CLIP relevancy score, we can create a buffer of point clouds in 3-dimensional space labeled with their object names.
\newpage

\subsection{Step Three}

Once we have the precise location of an object mapped over time, we then use this data to perform the Expectation-Maximization algorithm to generate a Gaussian Mixture Model.

The EM randomly initializes the means and covariance matrices of a specific number of Gaussian distributions. The algorithm then computes the posterior probabilities of each of the points. Once given these posterior probabilities, the means and the covariance matrices for the specified number of Gaussian distributions can be recalculated. This is done until convergence is reached.

\begin{figure}[htp]
    \centering
    \includegraphics[width=12cm]{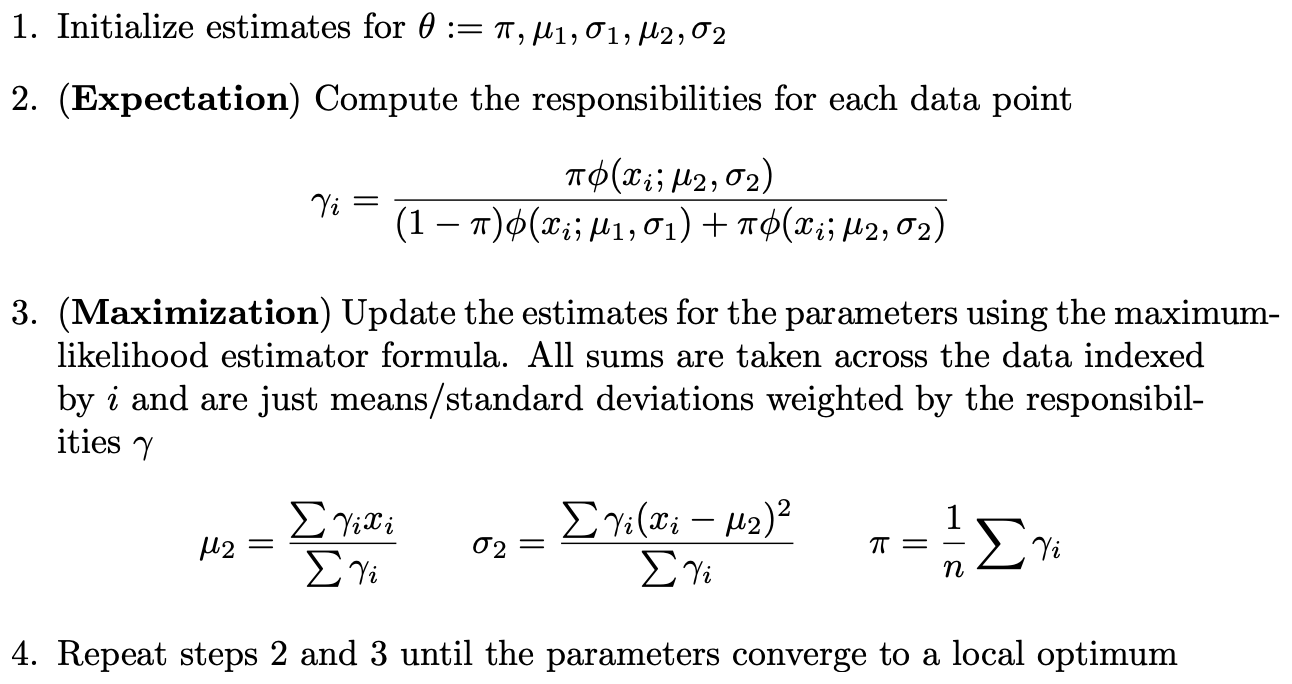}
    \caption{EM algorithm for two cluster centers}
    \label{fig:galaxy}
\end{figure}

A core concern for the EM algorithm is how to pick the correct number of clusters. We see two ways of determining this. One is to use conventional unsupervised learning approaches such as the Minimization of Bayesian Information Criterion.
$$BIC = k \cdot ln(n) -2ln(\hat{L})$$
Where $\hat{L} = p(x | \hat{\theta}, M)$ is the likelihood function of the model $M$, $x$ is the observed data, $\hat{\theta}$ are the observed parameter values, $\hat{k}$ is the number of individual Gaussians included in the computation, $n$ is the number of data points in $x$. 

These criteria exists to ensure that the chosen model doesn't over-fit an excessive number of Gaussian distributions to the model and fits the correct number of Gaussian distributions to the model. We utilized this approach in our code. 

Another approach might be to use LLMs to generate a number of clusters one might expect. For instance the number of different places in the household a LLM might return when queried "where is my wallet" could be a strong baseline for the number of clusters to expect. However we did not implement this approach.

\subsection{Step Four}
Upon providing a new query to the robot, the robot utilizes its saved history of an object's past locations generated above to sample a point from the distribution and then search for the object in that position. 

We ran simulations to see how much faster this approach was in comaprison to a naive randomized search. The overwhelming result was that it was much easier to find the object once you used the learned history of where the object has been historically placed.
\newpage

\section{Tests for Robustness}

\subsection{Localization of hidden objects}
We tested CLIP’s effectiveness in differentiating objects in different receptacles, similarly colored objects, and varied prepositions. Here are a few tests for robustness:
\begin{itemize}
  \item “potato in cabinet” and “bowl in cabinet” where the potato is placed inside the bowl resulted in a relevancy heat map that highlighted the area around the potato, but not the potato accurately when queried for the bowl
  \item “lettuce in fridge” and “tomato in fridge” clearly differentiated the two object locations because red and green are contrasting colors
  \item “bread above cabinet” demonstrated that the word above was correctly interpreted
  \item “bottle on countertop” was the first instance where the object wasn’t correctly localized because of its transparent nature
  \item “tomato in cabinet” was used to query an object that did not exist in the named receptacle, since CLIP highlights the object with the closest relevancy; this proved an important point about how the object’s receptacle container does not constrain the search with our current model
  \item “fork in open cabinet” as opposed to “fork in closed cabinet” queried on the most recent environment state once again proves that an identifier before the receptacle object doesn’t constrain search
\end{itemize}

\section{Search space optimization results:}

We assessed the robot's effectiveness in determining the correct cluster in which an object is located when given a varied number of training data examples.

We envisioned the task of searching for an item that has a ground truth distribution among locations. Consider there are three central cluster locations in which the object might be located. These locations contain an inherent probability distribution that is non-uniform. In such a situation, a random search would perform poorly.

\subsection{Procedure:}

A training data example consists of choosing a central cluster location at which the object will be spawned. We also add gaussian noise to mimic how humans might place an object in the house. For instance, if you place your keys on the kitchen table, they may not always be in a discrete position but instead are clustered around a central location.

The model is then trained on this data with a varying amount of sampled training examples.

We then test the model by spawning 100,000 random samples and testing how many times the model correctly finds the right location the first time. We evaluate a correct location by determining whether the determined location for the object is within 0.3 meters of the object's true location.

We benchmark the model's performance by comparing it to how a random search on the known cluster locations would perform.

\newpage
\subsection{Results and measurements:}
We ran the model with differing initial distributions among three locations of the object with corresponding probability distributions. 

\underline{\textbf{Distribution One:}} Consider the following asymmetric distribution over the three clusters: $Pr(\textrm{located in cluster 1})=0.7, Pr(\textrm{located in cluster 2})=0.2, Pr(\textrm{located in cluster 3})=0.1$ 
\begin{figure}[htp]
    \centering
    \includegraphics[scale=0.40]{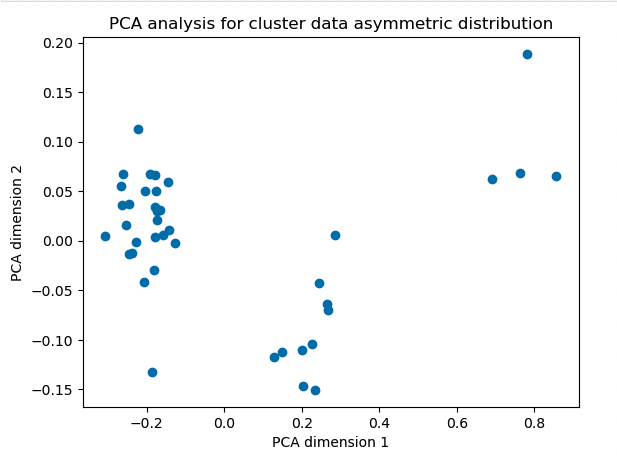}
    \caption{Distribution of 3-d location of tomato in asymmetric case projected onto 2-dimensions}
    \label{fig:galaxy}
\end{figure}

\begin{figure}[htp]
    \centering
    \includegraphics[scale=0.40]{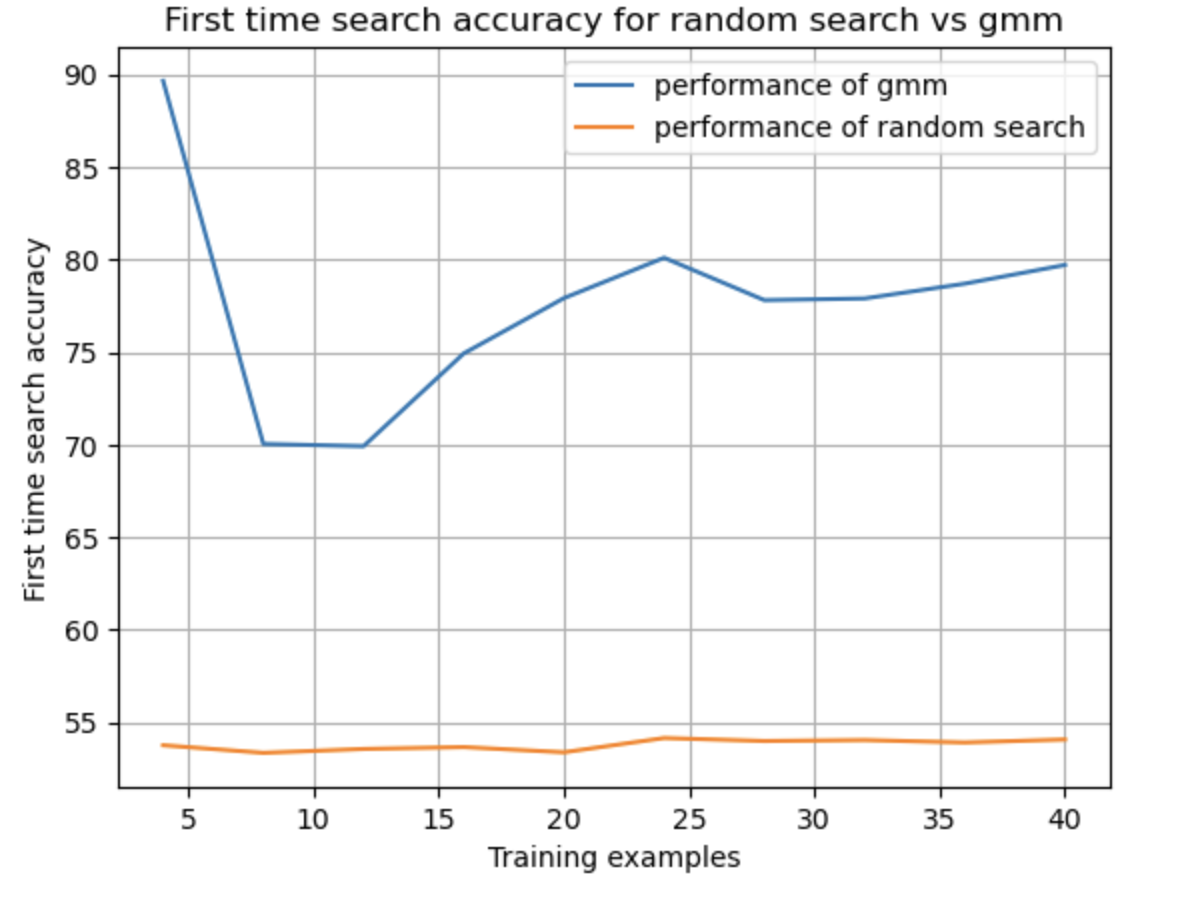}
    \caption{First-time search accuracy increases as compared to a brute force baseline }
    \label{fig:galaxy}
\end{figure}

\underline{\textbf{Distribution Two:}} Consider the following uniform distribution over the three clusters $Pr(\textrm{located in cluster 1})=\frac{1}{3}, Pr(\textrm{located in cluster 2})=\frac{1}{3}, Pr(\textrm{located in cluster 3})=\frac{1}{3}$ 

\begin{figure}[htp]
    \centering
    \includegraphics[scale=0.40]{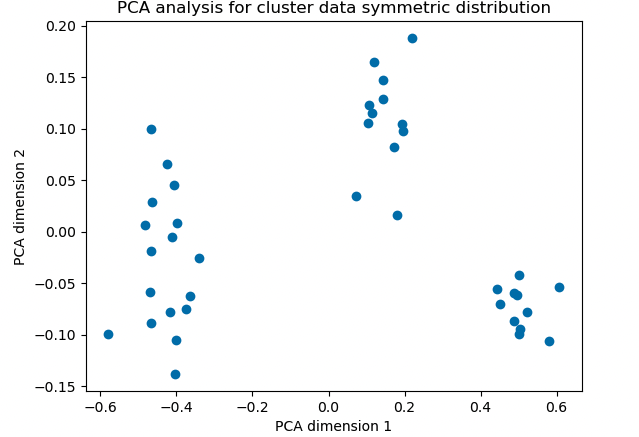}
    \caption{Distribution of 3-d location of tomato in symmetric case projected onto 2-dimensions}
    \label{fig:galaxy}
\end{figure}

\begin{figure}[htp]
    \centering
    \includegraphics[scale=0.40]{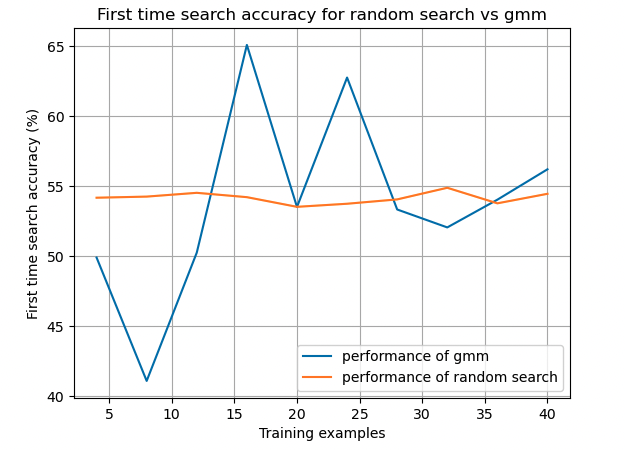}
    \caption{In the case where the prior distributions are the same, learning a GMM doesn't provide an advantage for search}
    \label{fig:galaxy}
\end{figure}

\newpage
\underline{\textbf{Analysis:}} Clearly in the first distribution the learned Gaussian model outperforms random search, as the distribution is asymmetric and the system learns to recognize this. However, when the distribution is symmetric the Gaussian model performs just as well as random search.

\newpage

\section{Extensions}
Determining the optimal number of clusters within our dataset of an object's past location is key to our method's effectiveness. This can be improved by queryingan  LLM  for possible locations an object could be based on it's understanding of words, which can be used to create the initial k clusters in the expectation-maximization function. The reason for his change is demonstrated in this example: Only the points on top of a table should be grouped together as opposed to the floor surrounding it because objects will usually never be intentionally placed there. Additionally, if there are two tables in an environment, they should be in distinct clusters since they represent completely different locations.

CLIP’s relevancy extractor does not work as a perfect baseline. In our testing image, there was a yellow sticker on the fridge and a green apple in a cabinet. Since both were similarly colored and round, they were both identified as highly relevant to the query “apple in the cabinet”. Our solution to this was to subtract the relevancy matrix for two images, where the cabinet with our desired apple was opened and closed, effectively eliminating the irrelevant pixels from the yellow sticker and preserving the pixels from the cabinet since the absolute value can be taken to still yield a positive relevancy value. This approach only works if the irrelevant pixels are highlighted in both images, which is not always true.

A more effective approach would be to take images from different angles with the same query “apple in the cabinet”. Since CLIP is given more information about the object’s shape, it is less likely to yield an incorrect result for a given query.

\begin{figure}[htp]
    \centering
    \includegraphics[scale=0.6]{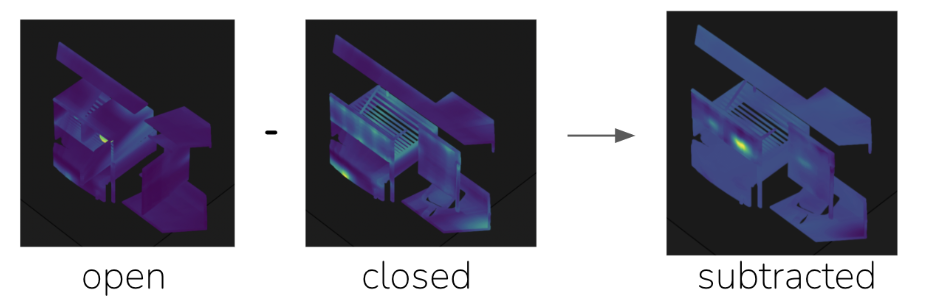}
    \caption{Resulting image does not identify irrelevant pixels }
    \label{fig:galaxy}
\end{figure}

A similar paper titled “Embodied Concept Learner”, or ECL, mimics how humans learn visual concepts and understand geometry and layout through active interaction with the environment. ECL could be beneficial for embodied instruction following (EIF), outperforming previous prior works on the well-known ALFRED benchmark. The learned concept of searching for objects can be reused for other downstream tasks, such as reasoning of object states and rearrangement.

Their approach simplifies the hard problem of making meaningful progress through tight integration between visual perception, language instruction, and robotic navigation and manipulation. To the best of our knowledge, no other benchmarks contain language instructions in an interactive 3D environment with visual observation and navigation. 

Furthermore, we developed this project from the perspective of using an egocentric-camera on a single robot. A single egocentric robot doesn't have full view of all rooms in a household and therefore will not be able to tell how objects are moving in rooms in which it is not located in. As a result it makes sense for this robot to learn about how an object is distributed. In our paper we have done this using a Gaussian mixture model. We see an extension to this paper where instead of working with Ego-Centric cameras, multiple cameras in a household could be combined to produce memorized trajectories of an object's location. This would provide an exact location of an object's whereabouts and would provide the best search optimization for an object.
\newpage

\section{Lessons Learned}

A key learning about the expectation maximization algorithm is that its performance was quite variable when we had a low amount of training data. For instance the algorithm performed better with 5 training samples then 10. This was because of the randomness associated with the algorithm and it's initial choice of distribution parameters. A lucky choice of distribution parameters would lead to an accurate modelling of the environment and better results. To combat this we would need several data samples - maybe in the order of 100s or 1000s. This may not be realistic. Further for objects that have multiple potential locations would need even more training data. Consider how many different places you keep your mobile phone - there could be 10 or 20 different locations. Learning a GMM on these locations would need substantial amounts of data. 

When a series of images are taken of objects in different configurations, we had to consider how unique objects can be tracked, ex: if the tomato in the fridge is no longer seen when the fridge is opened a second time, but a new tomato appears in the cabinet it should not be assumed that both instances are the same object. After working within a simulation environment for our entire project, we realized that objects do not just teleport to new locations as we wrote in our code, but instead they are manually transported. This means you can track the trajectory of an object even if some parts of its motion are obstructed from view. Using the same example, if the user queried the robot for “the tomato from the fridge” and not just “a tomato”, only if the same tomato from the fridge was seen moving into the cabinet, it will be identified.

The simulation environment we chose to use is roboTHOR but since we were running the semantic abstraction model using Google cloud’s GPU credits, the operating system on the virtual machine could not run the Unity window simultaneously. This required us to run robothor simulation on our local computers, save all camera metadata and object coordinate matrices in a .pkl file, and upload the file into the virtual machine where the model was running. Since the semantic abstraction model and simulation environment could not communicate with each other directly, it prevented us from making real-time decisions like opening the identified cabinet or moving to the desired location based on what was detected in the robot’s environment.

\section*{References}

\href{https://arxiv.org/pdf/2207.11514.pdf}{Semantic Abstraction: Open-World 3D Scene Understanding from 2D Vision-Language Models}

\href{https://openreview.net/pdf?id=yPJ9A0GWLg0}{Embodied Concept Learner: Self-supervised Learning of Concepts and Mapping through Instruction Following}

\small{}
\end{document}